\documentclass[10pt,twocolumn,letterpaper]{article}

\usepackage[pagenumbers]{cvpr} 
\usepackage{times}
\usepackage{helvet}
\usepackage{courier}
\usepackage{xcolor}
\usepackage{algorithm}
\usepackage{algorithmic}

\usepackage[table]{xcolor} 
\usepackage{booktabs}      
\usepackage{amsmath,amssymb,amsfonts}
\usepackage{subcaption} 
\usepackage{multirow}
\usepackage{siunitx}
\definecolor{cvprblue}{rgb}{0.21,0.49,0.74}
\usepackage[pagebackref,breaklinks,colorlinks,allcolors=cvprblue]{hyperref}

\def\paperID{281} 
\def\confName{CVPR}
\def\confYear{2026}

\title{ Histomorphology-Guided Prototypical Multi-Instance Learning \\for Breast Cancer WSI Classification}

\author{
Baizhi Wang\\
School of Biomedical Engineering, USTC\\
Center for Medical Imaging, Robotics, Analytic Computing \& Learning (MIRACLE), USTC\\
{\tt\small wbz@mail.ustc.edu.cn} 
\and
Rui Yan\thanks{Corresponding Author}\\
School of Biomedical Engineering, USTC\\
Center for Medical Imaging, Robotics, Analytic Computing \& Learning (MIRACLE), USTC\\
{\tt\small yanrui@ustc.edu.cn}
\and
Wenxin Ma\\
School of Biomedical Engineering, USTC\\
Center for Medical Imaging, Robotics, Analytic Computing \& Learning (MIRACLE), USTC\\
{\tt\small }
\and
Xu Zhang\\
School of Biomedical Engineering, USTC\\
Center for Medical Imaging, Robotics, Analytic Computing \& Learning (MIRACLE), USTC\\
{\tt\small }
\and
Yuhao Wang\\
School of Biomedical Engineering, USTC\\
Center for Medical Imaging, Robotics, Analytic Computing \& Learning (MIRACLE), USTC\\
{\tt\small }
\and
Xiaolong Li\\
School of Biomedical Engineering, USTC\\
Center for Medical Imaging, Robotics, Analytic Computing \& Learning (MIRACLE), USTC\\
{\tt\small }
\and
Yunjie Gu\\
School of Biomedical Engineering, USTC\\
Center for Medical Imaging, Robotics, Analytic Computing \& Learning (MIRACLE), USTC\\
{\tt\small }
\and
Zihang Jiang\\
School of Biomedical Engineering, USTC\\
Center for Medical Imaging, Robotics, Analytic Computing \& Learning (MIRACLE), USTC\\
{\tt\small }
\and
S.~Kevin Zhou\thanks{Corresponding Author}\\
School of Biomedical Engineering, USTC\\
Center for Medical Imaging, Robotics, Analytic Computing \& Learning (MIRACLE), USTC\\
Key Laboratory of Intelligent Information Processing, ICT, CAS\\
State Key Laboratory of Precision and Intelligent Chemistry, USTC\\
{\tt\small s.kevin.zhou@gmail.com}
}

\begin{document}
\maketitle

\begin{abstract}
Histomorphology is crucial in cancer diagnosis. However, existing whole slide image (WSI) classification methods struggle to effectively incorporate histomorphology information, limiting their ability to capture key pathological features. Particularly when the number of instances within a bag is large and their features are complex, it becomes challenging to accurately identify instances decisive for the bag label, making these methods prone to interference from ambiguous instances. To address this limitation, we propose a novel Histomorphology-Guided Prototypical Multi-Instance Learning (HGPMIL) framework that explicitly learns histomorphology-guided prototypical representations by incorporating tumor cellularity, cellular morphology, and tissue architecture. Specifically, our approach consists of three key components: (1) estimating the importance of tumor-related histomorphology information at patch-level based on medical prior knowledge; (2) generating representative prototypes through histomorphology-prototypical clustering; and (3) enabling WSI classification through histomorphology-guided prototypical aggregation.
HGPMIL adjusts the decision boundary by incorporating histomorphological importance to reduce instance label uncertainty, thereby reversely optimizing the bag-level boundary. Experimental results demonstrate its effectiveness, achieving high diagnostic accuracy for molecular subtyping, cancer subtyping and survival analysis. The code will be available.

\end{abstract}




\section{Introduction}

\begin{figure}[t]
    \centering
    \includegraphics[width=1\linewidth]{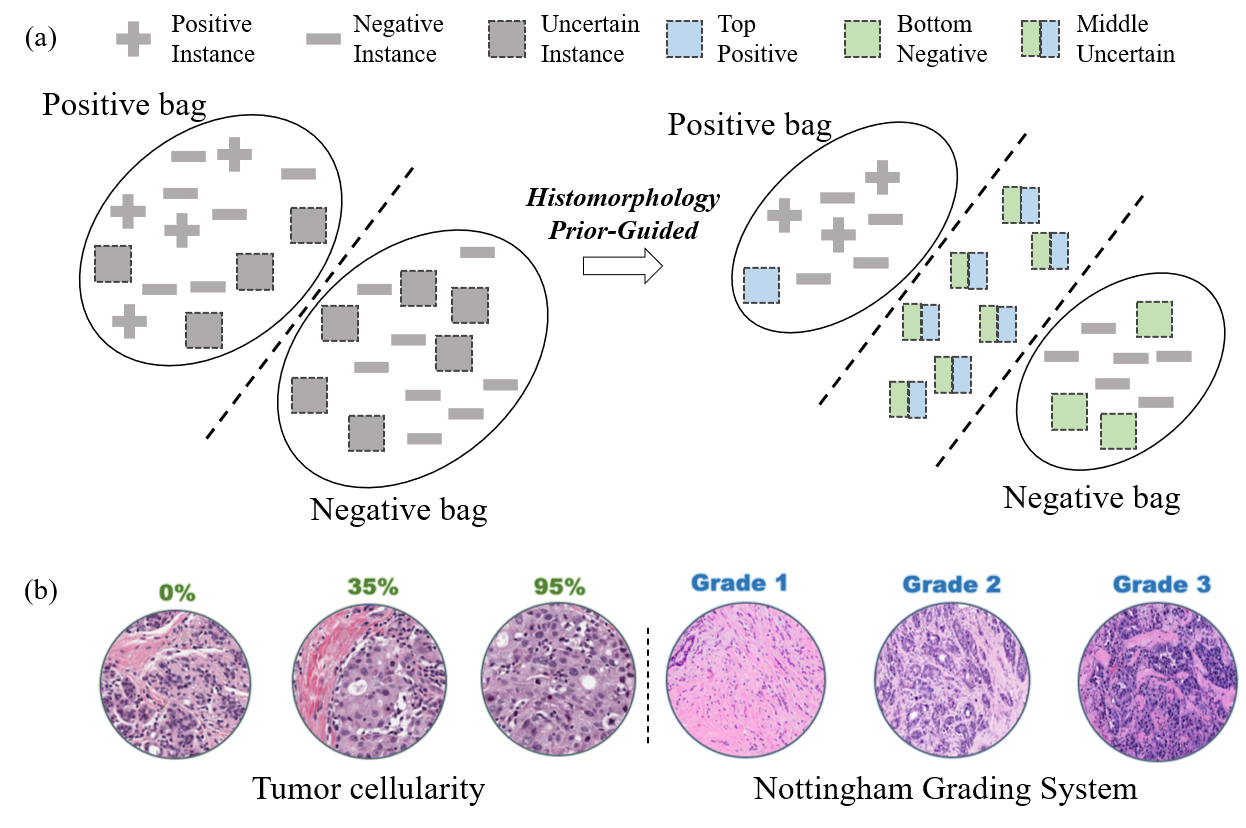}
    \caption{(a) Decision boundaries of MIL and HGPMIL. \textbf{Left}: MIL determines the decision boundary based on all patches equally. \textbf{Right}: HGPMIL adjusts the decision boundary by incorporating histomorphological importance to reduce instance label uncertainty, thereby reversely optimizing the bag-level boundary. (b) Examples illustrating histomorphology features characterized by tumor cellularity and the Nottingham Grading System.}
    \label{fig:example}
\end{figure}

Breast cancer is the most prevalent and lethal malignancy among women~\cite{nolan2023deciphering, blackadar2016historical}. Histomorphology features are critical for breast cancer~\cite{telloni2017tumor} diagnosis, grading, and prognosis. Histomorphology is typically evaluated based on cellular morphology and tissue architecture, as formalized in the Nottingham Grading System (NGS) \cite{elston1991pathological}, which scores nuclear atypia, mitotic count, and tubule formation, with these factors collectively determining the tumor grade (I–III); higher grades indicate poorer differentiation and more aggressive behavior, as illustrated in Fig. \ref{fig:example}(b). Meanwhile, tumor cellularity, defined as the percentage of malignant cells within the tumor bed, is another key indicator of tumor burden and is used to assess treatment response, especially after neoadjuvant therapy~\cite{rajan2004change}. A substantial reduction in cellularity correlates with favorable outcomes~\cite{heid2017co}. 


Computational pathology has significantly advanced whole slide image (WSI) analysis through deep learning~\cite{campanella2019clinical, chen2020deep,mobadersany2018predicting}, with most approaches relying on the weakly supervised learning paradigm known as multi-instance learning (MIL) ~\cite{ilse2018attention, courtiol2018classification}. However, existing computational models often fail to explicitly incorporate tumor tissue architecture and cellularity into decision making process, limiting their ability to capture key pathological features. Particularly when the number of instances (i.e., patches) within a bag (i.e., WSI) is large and their features are complex, it becomes challenging to accurately identify instances decisive for the bag label, making these methods prone to interference from ambiguous instances. 

To overcome these limitations, prototype learning has emerged as an effective approach for improving model performance and interpretability in computational pathology~\cite{hou2016patch,rymarczyk2022protomil,pmlryang24d,dingproto}. Prototypes, as learned representative feature centroids, allow the model to focus on biologically meaningful tumor regions, such as those with high cellularity or architecture~\cite{pan2023human,yang2023tpmil,yu2023prototypical}. By learning prototypes that represent the idealized features of tumor regions, the model can better differentiate between diagnostically relevant and irrelevant areas~\cite{struski2023promil, vu2023handcrafted, yu2023prototypical}. This enhances the decision boundary by reducing overlap between classes and ensuring that the model pays more attention to regions that are more informative and unambiguous for classification~\cite{song2024morphological}. 

In this paper, we propose \textbf{H}istomorphology-\textbf{G}uided \textbf{P}rototypical \textbf{M}ulti-\textbf{I}nstance \textbf{L}earning (HGPMIL), a  simple yet effective framework that learns histomorphological prototypes to guide instance aggregation, enabling more robust WSI classification. As shown in Fig.~\ref{fig:example}(a), HGPMIL achieves a more discriminative decision boundary than traditional MIL by leveraging histomorphological knowledge, leading to improved separation between positive and negative bags. Extensive experimental results demonstrate a strong correlation between utilizing tumor histomorphology and improving the accuracy of MIL-based WSI classification. The contributions of this paper can be summarized as follows:
\begin{itemize}
    \item We explicitly integrate histomorphological knowledge into WSI classification by learning prototypical representations guided by medical priors. This enables the model to capture critical pathological patterns while filtering out ambiguous instances, which are essential for MIL-based classification.
    \item We propose HGPMIL, a novel and adaptable histomorphology-guided prototypical multi-instance learning framework, designed to enhance prototypical aggregation through three key components: histomorphology importance estimation network, histomorphology-prototypical clustering, and histomorphology-guided prototypical aggregation.
    \item Extensive experiments demonstrate that HGPMIL is highly flexible and broadly applicable across various MIL models and downstream tasks, consistently leading to significant performance improvements.
\end{itemize} 

\section{Related Work}

\subsection{Attention-based and Knowledge-guided MIL}
MIL has become a standard weakly supervised learning paradigm in computational pathology, especially when detailed cell- or region-level annotations are unavailable~\cite{skrede2020deep}. In MIL, a WSI is treated as a bag of instances (typically patches), and the goal is to infer a slide-level label by aggregating instance-level information~\cite{javed2022additive,thandiackal2022differentiable}. Attention-based methods, such as ABMIL~\cite{ilse2018attention}, CLAM~\cite{lu2021data}, and TransMIL~\cite{shao2021transmil}, have shown strong performance by learning attention weights that emphasize informative patches. These models improve both accuracy and interpretability by enabling instance-level localization of diagnostically relevant regions. More recently, HDMIL~\cite{yan2023histopathological} extends this line of work with a hierarchical attention mechanism to capture both local and global context. However, while effective, most of these models rely solely on data-driven attention and do not explicitly incorporate medical priors.

\subsection{Efficient and Long-sequence MIL}
Processing WSIs with thousands of patches introduces computational and modeling challenges~\cite{chen2022scaling} . Transformer-based MIL methods like TransMIL and DTFD-MIL~\cite{zhang2022dtfd} address this by modeling long-range dependencies. Meanwhile, state-space sequence models, such as S4MIL~\cite{fillioux2023structured} and MambaMIL~\cite{yang2024mambamil}, provide more scalable alternatives with linear-time complexity and strong capacity for long-context modeling. These methods achieve competitive performance while reducing computational costs. However, they typically treat all instances equally and overlook histopathological relevance at the patch level.

Despite these advances, existing MIL frameworks \cite{WAQAS2024123893} often process large numbers of redundant  patches and lack mechanisms to incorporate domain knowledge into patch evaluation. This limits efficiency and clinical interpretability. To address these challenges, we propose HGPMIL, a plug-and-play framework that integrates tumor tissue architecture and cellularity into MIL process. By guiding patch importance estimation and aggregation with histomorphological priors, HGPMIL enhances both diagnostic accuracy and biological relevance in WSI-level classification.

\section{Methods}

\begin{figure*}[t]
    \centering
    \includegraphics[width=1\linewidth]{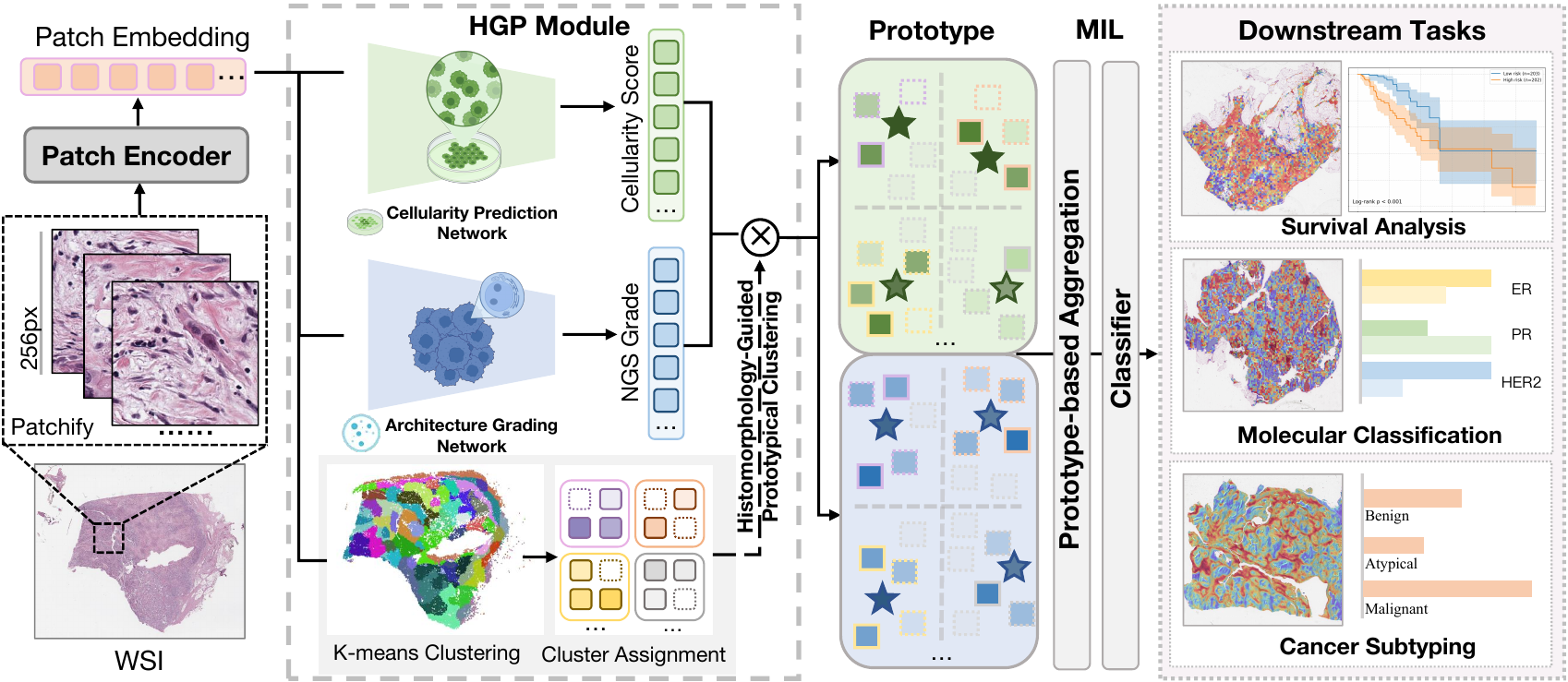}
    \caption{Overview of HGPMIL. Given a set of patches cropped from a slide, we sequentially utilize Patch Encoder, Cellularity Prediction Network, Architecture Grading Network, Histomorphology-Guided Prototypical Clustering, Histomorphology-Guided Prototypical Aggregation, and Multi-Instance Learning for WSI analysis of different downstream tasks. The shade of color within the rectangles reflects the importance of sorted tumor-related histomorphological information: the darker the color, the higher the importance, while gray ones represent middle-ranked uncertain instances.}
    \label{fig:framework}
\end{figure*}


The overview of HGPMIL is illustrated in Fig.~\ref{fig:framework}. In addition to traditional MIL components, i.e. patch encoder, feature aggregation module, and classifier, HGPMIL contains three novel designs: Histmorphology-centric importance estimation network, Histomorphology-prototypical clustering and Histomorphology-guided prototypical aggregation.

To begin with, WSIs are divided into non-overlapping 256 × 256 patches at 20× magnification, which are subsequently encoded into patch embeddings:
\begin{equation}
    X=\text{PatchEncoder}(P),
\end{equation}
where $P \in \mathbb{R}^{N\times 3\times 256 \times 256}$ is the input patch, and $X \in \mathbb{R}^{N\times D}$ is the extracted patch embedding.

\subsection{Histomorphology-Centric importance estimation network}
To evaluate the importance of tumor histomorphological information at the patch level based on medical prior knowledge, our method employs two auxiliary networks to predict tumor cellularity and tissue architecture of each instance. Notably, our proposed method exhibits generality, and the assessment of importance can also be implemented via other approaches, such as text-based descriptions.

\subsubsection{Cellularity prediction network (CPN)}  
To estimate instance-level tumor cellularity, we introduce the Cellularity Prediction Network (CPN), a lightweight neural network that predicts the cellularity score for each instance. The CPN is trained using the BreastPathQ dataset~\cite{martel2019assessment}, which contains high-quality, instance-level annotations for tumor cellularity assessment. 
Formally, the predicted cellularity score for a given instance \(X\) is:
\begin{equation}
    C_N = CPN(X),
\end{equation}
where $CPN(\cdot)$ denotes the trained CPN model, and $C_N\in \mathbb{R}^{N}$ represents the predicted instance-level cellularity score. This constraint refines the model’s decision boundary, making it precise by emphasizing areas with high tumor density.


\subsubsection{Architecture grading network (AGN)}  
Similarly, to estimate instance-level tumor tissue architecture, we design the Architecture Grading Network (AGN), which predicts the NGS score for each instance. Trained on the MITOS-ATYPIA-14~\cite{Roux2014} dataset, the AGN provides detailed annotations for histological grading. For a given input instance feature \(X\), the predicted tumor tissue architecture grade $A_N$ is estimated as: 
\begin{equation}
    A_N = AGN(X),
\end{equation}
where $AGN(\cdot)$ is the trained AGN model, and $A_N\in \mathbb{R}^{N}$ denotes the predicted instance-level tumor tissue architecture. 
By incorporating architecture grade as a constraint, the decision boundary becomes more finely tuned, improving the model’s ability to distinguish between subtle morphological differences that are critical for accurate diagnosis.

\subsection{Histomorphology-prototypical Clustering}
Due to the significant redundancy present in WSIs, we apply histomorphology-prototypical clustering to generate prototypes. In traditional clustering, patches are grouped based on their proximity in the feature space. However, our method integrates histomorphological knowledge to enhance the clustering process, making it more aligned with biologically meaningful tumor features.
For each instance, we concatenate the patch embedding \( X_i \) with the predicted \textit{tumor cellularity ($C_i$)} and \textit{architecture grade ($A_i$)}, forming an extended feature vector:

\begin{equation}
X_i^{'} = [X_i; C_i; A_i].
\end{equation}
Here, \( C_i \) and \( A_i \) are the predicted tumor cellularity and architecture grade for instance \( i \), respectively.

After extending the feature space, we apply K-means clustering to partition the instances into \( K \) clusters. In each cluster \( R_k \), the cluster center \( P_k \) is computed as the prototype representing the typical features of the instances within that cluster:
\begin{equation}
P_k = \frac{1}{N_k} \sum_{i \in R_k} X_i^{'},
\end{equation}
where \( P_k \) is the prototype for the \( k \)-th cluster, and \( N_k \) is the number of instances in cluster \( R_k \).
The set of instances assigned to the \( k \)-th cluster is denoted as \( S_k \), where each instance \( X_i \) is part of the cluster centered around the prototype \( P_k \):
\begin{equation}
S_k = \{ X_i \ | \ X_i \in R_k \}.
\end{equation}

\subsection{Histomorphology-guided prototypical aggregation}
We generate two types of prototypes based on the predicted tumor cellularity and architecture grade. These prototypes represent the positive and negative regions within each cluster, emphasizing areas of high and low tumor cellularity or architecture grade, respectively. 

Specifically, we generate two masks that correspond to the top $R\%$ and bottom $R\%$ of instances within each cluster, based on their tumor cellularity scores. The top $R\%$ are used to generate positive prototypes that highlight regions with high cellularity, while the bottom $R\%$ represent negative prototypes for regions with low cellularity.
The positive and negative prototypes are represented as:
\begin{align}
    X_{C,k}^{pos} &= Norm((M_{C,k}^{pos}\odot C_k)S_k),\\
    X_{C,k}^{neg} &= Norm((M_{C,k}^{neg}\odot C_k)S_k),
\end{align}
where $M_{C,k}^{pos},M_{C,k}^{neg} \in \mathbb{R}^{N_k}$ mean the positive and negative cellularity instance masks of the $k$-th cluster, $C_k \in \mathbb{R}^{N_k}$ means the corresponding tumor cellularity, and $S_k \in \mathbb{R}^{N_k \times D}$ means the set of instance features.

These positive and negative prototypes provide more discriminative representations of the tumor instances. By incorporating these prototypes, we improve the model's ability to focus on the most diagnostically relevant areas, while filtering out ambiguous instances.

Similarly, we generate positive and negative prototypes for each cluster based on architecture grade. This allows the model to focus on patches with distinguishing structural characteristics that are important for the tumor's diagnostic and prognostic interpretation.
The positive and negative architecture prototypes are represented as:
\begin{align}
    X_{A,k}^{pos} &= Norm((M_{A,k}^{pos}\odot A_k)S_k),\\
    X_{A,k}^{neg} &= Norm((M_{A,k}^{neg}\odot A_k)S_k),
\end{align}
where $M_{A,k}^{pos},M_{A,k}^{neg} \in \mathbb{R}^{N_k}$ mean the positive and negative architecture instance masks of the $k$-th cluster, $A_k\in \mathbb{R}^{N_k}$ means the corresponding tumor tissue architecture grade, and $S_k \in \mathbb{R}^{N_k \times D}$ means the set of instance features. This features allow the model to distinguish between regions with significant structural differences.

Consequently, we generate four distinct prototypes for each cluster:
\begin{align}
\mathcal{X}_k= \{X_{C,k}^{pos},X_{C,k}^{neg},X_{A,k}^{pos},X_{A,k}^{neg}\}.
\end{align}
These prototypes serve as the representative features for each cluster and guide the multi-instance learning.

\subsection{Multi-instance learning with HGP} 
The \textbf{H}istomorphology-\textbf{G}uided \textbf{P}rotorype (HGP) design can be seamlessly integrated into various MIL frameworks.
In MIL, the training dataset consists of multiple clusters, each assigned a label, while the individual instances within a cluster remain unlabeled. The cluster’s label is determined by the presence of positive instances: if at least one instance in a cluster is positive, the entire cluster is labeled as positive; if all instances are negative, the cluster is classified as negative.  
Given an instance $x_d$ and corresponding label $y_d \in \{0,1\}$ in a cluster $B=\left\{\left(x_1, y_1\right), \ldots,\left(x_D, y_D\right)\right\}$, the label of $B$ is given by:
\begin{equation}
Y=\prod_{d=1}^D y_d=\left\{\begin{array}{l}
0,\quad \forall \ y_d=0, \\
1,\quad \exists \  y_d=1.
\end{array}\right. 
\end{equation}
Subsequently, the WSI-level representation is fed into different classifiers to perform different downstream tasks.

\section{Experiments}
\subsection{Datasets and evaluation metrics}
We focus on three tasks: molecular subtyping, cancer subtyping and survival analysis.
For molecular subtyping, we perform experiments on estrogen receptor (ER), progesterone receptor (PR), and human epidermal growth factor receptor 2 (HER2) classification using datasets from TCGA-BRCA~\cite{tomczak2015review} and BCNB~\cite{xu2021predicting}. For cancer subtyping, We perform extensive experiments on two public datasets: BRACS~\cite{brancati2022bracs} and TCGA-BRCA. For survival analysis, We perform experiments on
TCGA-BRCA dataset, and the univariate
KM-estimation (with Log-rank p-values) is used as the metric \cite{ruiyanpamt}.
To mitigate the impact of data partition, we employ 5-fold cross-validation in subtyping task. For performance evaluation, we adopt the Area Under the ROC Curve (AUC) and Accuracy (ACC) metrics, along with their standard deviation (std), following standard practice. These metrics provide a robust assessment while reducing sensitivity to class imbalance.

\subsection{Implementation details}

To evaluate the effectiveness of HGP design, we compare it against several state-of-the-art methods, including Max Pooling, attention-based multiple instance learning (ABMIL \cite{ilse2018attention}) and its variants CLAM-SB \cite{lu2021data}, as well as Transformer-based approaches like TransMIL \cite{shao2021transmil}. Additionally, we evaluate SSM-based S4MIL \cite{fillioux2023structured} and the Mamba-based MambaMIL \cite{yang2024mambamil}. To ensure fair comparisons, the patch embeddings are extracted using the Gigapath \cite{xu2024whole} patch encoder, which adopts the ViT-G/14 architecture and is pre-trained via DINOv2 \cite{oquab2023dinov2} method. We adopt the same data preprocessing pipeline as Gigapath. We select $K$ = 50 as the number of clusters \cite{ruiyanpamt} and $R$ = 50\% as the select ratio to ensure the best clustering performance.

\subsection{Comparison results}
\subsubsection{Molecular subtyping}

Molecular subtyping is essential for guiding breast cancer treatment, as ER, PR, and HER2 status directly influence therapeutic decisions. Accurate prediction of these markers from WSIs can reduce reliance on manual immunohistochemistry assays and enable scalable, consistent diagnostic workflows. Advancing this task holds significant clinical and practical value in computational pathology.

\begin{table*}[!htbp]\small
\centering
\caption{Molecular Subtyping Results on TCGA-BRCA  dataset.}
\renewcommand{\arraystretch}{1.25}
\newcommand{\SPM}[1]{\scalebox{1}{\ensuremath{\pm #1}}}

\begin{tabular}{l|cc|cc|cc|cc}
\hline
\rowcolor{gray!15}
\multicolumn{9}{c}{\textbf{TCGA-BRCA}} \\

\hline

& \multicolumn{2}{c|}{\textbf{ER}}
& \multicolumn{2}{c|}{\textbf{PR}}
& \multicolumn{2}{c|}{\textbf{HER2}}
& \multicolumn{2}{c}{\textbf{Mean}} \\

 \cmidrule(lr){2-3}\cmidrule(lr){4-5}\cmidrule(lr){6-7}\cmidrule(lr){8-9}
\textbf{Method} 
& \textbf{AUC} & \textbf{ACC} 
& \textbf{AUC} & \textbf{ACC} 
& \textbf{AUC} & \textbf{ACC} 
& \textbf{AUC} & \textbf{ACC} \\
\hline
Max Pooling
& 0.811\SPM{0.070} & 0.802\SPM{0.050}
& 0.828\SPM{0.055} & 0.776\SPM{0.063}
& 0.718\SPM{0.058} & 0.819\SPM{0.027}
& 0.785 & 0.799 \\

\rowcolor{gray!10}
+HGP(ours)
& \textbf{0.861}\SPM{0.041} & \textbf{0.841}\SPM{0.060}
& \textbf{0.830}\SPM{0.050} & \textbf{0.785}\SPM{0.059}
& \textbf{0.759}\SPM{0.041} & \textbf{0.823}\SPM{0.048}
& \textbf{0.816} & \textbf{0.816} \\
\hline
TransMIL
& 0.853\SPM{0.041} & \textbf{0.806}\SPM{0.062}
& 0.794\SPM{0.051} & 0.750\SPM{0.059}
& \textbf{0.710}\SPM{0.089} & 0.830\SPM{0.048}
& 0.785 & 0.795 \\

\rowcolor{gray!10}
+HGP(ours)
& \textbf{0.863}\SPM{0.062} & \textbf{0.806}\SPM{0.052}
& \textbf{0.838}\SPM{0.056} & \textbf{0.760}\SPM{0.050}
& 0.683\SPM{0.093} & \textbf{0.832}\SPM{0.065}
& \textbf{0.794} & \textbf{0.799} \\
\hline
MambaMIL
& 0.867\SPM{0.035} & 0.828\SPM{0.047}
& 0.829\SPM{0.070} & 0.783\SPM{0.073}
& 0.749\SPM{0.027} & 0.839\SPM{0.060}
& 0.815 & 0.816 \\

\rowcolor{gray!10}
+HGP(ours)
& \textbf{0.881}\SPM{0.030} & \textbf{0.830}\SPM{0.038}
& \textbf{0.866}\SPM{0.082} & \textbf{0.808}\SPM{0.060}
& \textbf{0.755}\SPM{0.072} & \textbf{0.846}\SPM{0.018}
& \textbf{0.834} & \textbf{0.828} \\
\hline
ABMIL
& 0.851\SPM{0.059} & 0.805\SPM{0.059}
& 0.843\SPM{0.067} & 0.793\SPM{0.095}
& 0.783\SPM{0.030} & \textbf{0.815}\SPM{0.024}
& 0.825 & 0.804 \\

\rowcolor{gray!10}
+HGP(ours)
& \textbf{0.859}\SPM{0.061} & \textbf{0.826}\SPM{0.091}
& \textbf{0.862}\SPM{0.055} & \textbf{0.812}\SPM{0.050}
& \textbf{0.797}\SPM{0.018} & \textbf{0.815}\SPM{0.021}
& \textbf{0.839} & \textbf{0.817} \\
\hline
CLAM-SB
& 0.870\SPM{0.055} & \textbf{0.843}\SPM{0.066}
& 0.854\SPM{0.065} & \textbf{0.814}\SPM{0.065}
& 0.747\SPM{0.038} & 0.799\SPM{0.039}
& 0.823 & 0.818 \\

\rowcolor{gray!10}
+HGP(ours)
& \textbf{0.870}\SPM{0.035} & 0.834\SPM{0.051}
& \textbf{0.865}\SPM{0.057} & 0.808\SPM{0.045}
& \textbf{0.767}\SPM{0.052} & \textbf{0.838}\SPM{0.065}
& \textbf{0.834} & \textbf{0.826} \\
\hline
S4MIL
& \textbf{0.871}\SPM{0.026} & \textbf{0.834}\SPM{0.040}
& 0.859\SPM{0.067} & 0.797\SPM{0.059}
& 0.772\SPM{0.055} & \textbf{0.844}\SPM{0.051}
& 0.834 & \textbf{0.825} \\

\rowcolor{gray!10}
+HGP(ours)
& 0.868\SPM{0.023} & 0.814\SPM{0.037}
& \textbf{0.865}\SPM{0.067} & \textbf{0.803}\SPM{0.076}
& \textbf{0.782}\SPM{0.022} & 0.836\SPM{0.040}
& \textbf{0.838} & 0.817 \\
\hline
\end{tabular}
\label{tab:TCGAmolecular}
\end{table*}

\begin{table*}[!htbp]\small
\centering
\caption{Molecular Subtyping Results on BCNB dataset.}
\renewcommand{\arraystretch}{1.25}
\newcommand{\SPM}[1]{\scalebox{1}{\ensuremath{\pm #1}}}

\begin{tabular}{l|cc|cc|cc|cc}
\hline
\rowcolor{gray!15}
\multicolumn{9}{c}{\textbf{BCNB}} \\
\hline
& \multicolumn{2}{c|}{\textbf{ER}}
& \multicolumn{2}{c|}{\textbf{PR}}
& \multicolumn{2}{c|}{\textbf{HER2}}
& \multicolumn{2}{c}{\textbf{Mean}} \\
 \cmidrule(lr){2-3}\cmidrule(lr){4-5}\cmidrule(lr){6-7}\cmidrule(lr){8-9}
\textbf{Method} 
& \textbf{AUC} & \textbf{ACC} 
& \textbf{AUC} & \textbf{ACC} 
& \textbf{AUC} & \textbf{ACC} 
& \textbf{AUC} & \textbf{ACC} \\
\hline

Max Pooling
& 0.893\SPM{0.055} & \textbf{0.852}\SPM{0.053}
& 0.805\SPM{0.043} & 0.775\SPM{0.033}
& 0.671\SPM{0.045} & 0.715\SPM{0.039}
& 0.789 & 0.780 \\

\rowcolor{gray!10}
+HGP(ours)
& \textbf{0.894}\SPM{0.046} & 0.839\SPM{0.047}
& \textbf{0.807}\SPM{0.037} & \textbf{0.792}\SPM{0.041}
& \textbf{0.700}\SPM{0.059} & \textbf{0.731}\SPM{0.064}
& \textbf{0.800} & \textbf{0.787} \\
\hline
TransMIL
& 0.859\SPM{0.040} & \textbf{0.859}\SPM{0.040}
& 0.757\SPM{0.025} & 0.677\SPM{0.108}
& 0.680\SPM{0.046} & 0.679\SPM{0.088}
& 0.765 & 0.738 \\

\rowcolor{gray!10}
+HGP(ours)
& \textbf{0.867}\SPM{0.035} & 0.828\SPM{0.020}
& \textbf{0.783}\SPM{0.020} & \textbf{0.777}\SPM{0.056}
& \textbf{0.684}\SPM{0.052} & \textbf{0.713}\SPM{0.041}
& \textbf{0.778} & \textbf{0.772} \\
\hline
MambaMIL
& 0.894\SPM{0.023} & 0.828\SPM{0.031}
& 0.783\SPM{0.026} & 0.779\SPM{0.039}
& 0.674\SPM{0.053} & 0.707\SPM{0.038}
& 0.783 & 0.771 \\

\rowcolor{gray!10}
+HGP(ours)
& \textbf{0.911}\SPM{0.044} & \textbf{0.858}\SPM{0.046}
& \textbf{0.816}\SPM{0.017} & \textbf{0.803}\SPM{0.027}
& \textbf{0.690}\SPM{0.058} & \textbf{0.713}\SPM{0.047}
& \textbf{0.805} & \textbf{0.791} \\
\hline
ABMIL
& 0.914\SPM{0.037} & 0.845\SPM{0.060}
& 0.821\SPM{0.025} & 0.777\SPM{0.039}
& 0.703\SPM{0.072} & \textbf{0.728}\SPM{0.036}
& 0.812 & 0.783 \\

\rowcolor{gray!10}
+HGP(ours)
& \textbf{0.916}\SPM{0.035} & \textbf{0.864}\SPM{0.043}
& \textbf{0.827}\SPM{0.017} & \textbf{0.785}\SPM{0.036}
& \textbf{0.736}\SPM{0.058} & \textbf{0.728}\SPM{0.043}
& \textbf{0.826} & \textbf{0.792} \\
\hline
CLAM-SB
& 0.899\SPM{0.053} & 0.850\SPM{0.055}
& \textbf{0.794}\SPM{0.040} & 0.783\SPM{0.044}
& 0.697\SPM{0.067} & 0.720\SPM{0.063}
& 0.796 & 0.784 \\

\rowcolor{gray!10}
+HGP(ours)
& \textbf{0.908}\SPM{0.035} & \textbf{0.854}\SPM{0.039}
& 0.792\SPM{0.045} & \textbf{0.790}\SPM{0.047}
& \textbf{0.709}\SPM{0.071} & \textbf{0.743}\SPM{0.043}
& \textbf{0.803} & \textbf{0.795} \\
\hline
S4MIL
& 0.898\SPM{0.037} & 0.847\SPM{0.027}
& \textbf{0.823}\SPM{0.022} & \textbf{0.794}\SPM{0.012}
& 0.708\SPM{0.046} & 0.735\SPM{0.027}
& 0.809 & 0.792 \\

\rowcolor{gray!10}
+HGP(ours)
& \textbf{0.901}\SPM{0.046} & \textbf{0.858}\SPM{0.053}
& \textbf{0.823}\SPM{0.033} & 0.790\SPM{0.039}
& \textbf{0.719}\SPM{0.055} & \textbf{0.737}\SPM{0.030}
& \textbf{0.814} & \textbf{0.795} \\
\hline

\end{tabular}
\label{tab:BCNBmolecular}
\end{table*}

HGP consistently improves molecular subtyping performance across diverse MIL baselines, demonstrating strong generalizability and robustness across cohorts and subtype targets. It yields gains in both AUC and ACC on TCGA-BRCA and BCNB datasets, with the most notable improvements seen in MambaMIL (+1.9\%/+1.2\% on TCGA-BRCA and +2.2\%/+2.0\% on BCNB). Similar trends are observed in ABMIL, CLAM-SB, and S4MIL, where HGP consistently enhances performance over their original variants. HGP effectively guides them to focus on histomorphology relevant regions, resulting in more discriminative feature representations. These results (Table~\ref{tab:TCGAmolecular} \& Table~\ref{tab:BCNBmolecular}) confirm that HGP is a broadly applicable and effective enhancement.

\subsubsection{Cancer subtyping}

Cancer subtyping is critical for determining prognosis and informing treatment strategies. Unlike binary classification tasks, subtyping often involves fine-grained distinctions between morphologically similar categories, making it more challenging and reliant on nuanced histological features. 

\begin{table*}[t]\small
\centering
\caption{Cancer Subtyping Results on BRACS and TCGA-BRCA datasets.}
\renewcommand{\arraystretch}{1.25}
\newcommand{\SPM}[1]{\scalebox{1}{\ensuremath{\pm #1}}}

\begin{tabular}{l|cc|cc|cc|cc}
\hline
\rowcolor{gray!15}
\multicolumn{9}{c}{\textbf{Cancer Subtyping}} \\
\hline
& \multicolumn{2}{c|}{\textbf{BRACS-7}}
& \multicolumn{2}{c|}{\textbf{BRACS-3}}
& \multicolumn{2}{c|}{\textbf{TCGA-BRCA-2}}
& \multicolumn{2}{c}{\textbf{Mean}} \\
 \cmidrule(lr){2-3}\cmidrule(lr){4-5}\cmidrule(lr){6-7}\cmidrule(lr){8-9}

\textbf{Method}& \textbf{AUC} & \textbf{ACC}
& \textbf{AUC} & \textbf{ACC}
& \textbf{AUC} & \textbf{ACC}
& \textbf{AUC} & \textbf{ACC} \\
\hline
Max Pooling 
& 0.784\SPM{0.031} & 0.505\SPM{0.041}
& 0.847\SPM{0.035} & 0.724\SPM{0.054}
& 0.873\SPM{0.050} & 0.845\SPM{0.034}
& 0.834 & 0.691 \\
\rowcolor{gray!10}
+HGP(ours)
& \textbf{0.808}\SPM{0.039} & \textbf{0.513}\SPM{0.052}
& \textbf{0.855}\SPM{0.031} & \textbf{0.742}\SPM{0.049}
& \textbf{0.878}\SPM{0.047} & \textbf{0.851}\SPM{0.057}
& \textbf{0.847} & \textbf{0.702} \\
\hline
TransMIL 
& 0.789\SPM{0.044} & 0.460\SPM{0.021}
& 0.875\SPM{0.045} & \textbf{0.734}\SPM{0.057}
& 0.883\SPM{0.033} & 0.851\SPM{0.028}
& 0.849 & 0.681 \\
\rowcolor{gray!10}
+HGP(ours)
& \textbf{0.798}\SPM{0.032} & \textbf{0.462}\SPM{0.054}
& \textbf{0.878}\SPM{0.026} & 0.732\SPM{0.045}
& \textbf{0.887}\SPM{0.042} & \textbf{0.856}\SPM{0.039}
& \textbf{0.854} & \textbf{0.683} \\
\hline
MambaMIL 
& \textbf{0.831}\SPM{0.035} & \textbf{0.513}\SPM{0.067}
& 0.888\SPM{0.043} & 0.721\SPM{0.025}
& \textbf{0.873}\SPM{0.071} & \textbf{0.825}\SPM{0.035}
& 0.864 & \textbf{0.686} \\
\rowcolor{gray!10}
+HGP(ours)
& 0.822\SPM{0.037} & 0.468\SPM{0.047}
& \textbf{0.901}\SPM{0.020} & \textbf{0.762}\SPM{0.047}
& 0.871\SPM{0.007} & 0.810\SPM{0.059}
& \textbf{0.865} & 0.680 \\
\hline
ABMIL 
& 0.835\SPM{0.037} & 0.520\SPM{0.063}
& 0.895\SPM{0.026} & 0.751\SPM{0.062}
& \textbf{0.901}\SPM{0.022} & 0.862\SPM{0.021}
& 0.877 & 0.711 \\
\rowcolor{gray!10}
+HGP(ours)
& \textbf{0.839}\SPM{0.030} & \textbf{0.524}\SPM{0.055}
& \textbf{0.899}\SPM{0.039} & \textbf{0.777}\SPM{0.084}
& \textbf{0.901}\SPM{0.018} & \textbf{0.876}\SPM{0.021}
& \textbf{0.879} & \textbf{0.725} \\
\hline
CLAM-sb 
& 0.840\SPM{0.040} & 0.543\SPM{0.043}
& 0.872\SPM{0.036} & \textbf{0.743}\SPM{0.047}
& \textbf{0.905}\SPM{0.009} & 0.870\SPM{0.018}
& 0.872 & 0.718 \\
\rowcolor{gray!10}
+HGP(ours) 
& \textbf{0.845}\SPM{0.028} & \textbf{0.554}\SPM{0.010}
& \textbf{0.902}\SPM{0.039} & \textbf{0.743}\SPM{0.049}
& \textbf{0.905}\SPM{0.013} & \textbf{0.872}\SPM{0.014}
& \textbf{0.884} & \textbf{0.723} \\
\hline
S4MIL 
& 0.827\SPM{0.031} & 0.520\SPM{0.059}
& \textbf{0.904}\SPM{0.024} & \textbf{0.773}\SPM{0.040}
& 0.895\SPM{0.018} & 0.860\SPM{0.030}
& 0.875 & 0.717 \\
\rowcolor{gray!10}
+HGP(ours)
& \textbf{0.844}\SPM{0.037} & \textbf{0.547}\SPM{0.055}
& 0.897\SPM{0.036} & 0.735\SPM{0.042}
& \textbf{0.899}\SPM{0.015} & \textbf{0.874}\SPM{0.020}
& \textbf{0.880} & \textbf{0.718} \\
\hline

\end{tabular}
\label{tab:cancer}
\end{table*}
HGP demonstrates consistent performance gains across most baselines on the BRACS and TCGA-BRCA datasets, validating its ability to enhance feature aggregation in multi-class settings. As summarized in Table~\ref{tab:cancer}, the largest improvements are observed in attention-based models such as ABMIL and CLAM-SB, with ABMIL achieving +1.2\% AUC and +1.4\% ACC gains. Even in strong baselines like S4MIL and TransMIL, HGP still leads to consistent boosts. These results suggest that HGP is particularly effective in guiding MILs to focus on histologically informative regions, thereby improving class separation in complex subtyping scenarios.

\subsubsection{Survival analysis}
\begin{figure}[t]
    \centering
    \includegraphics[width=1\linewidth]{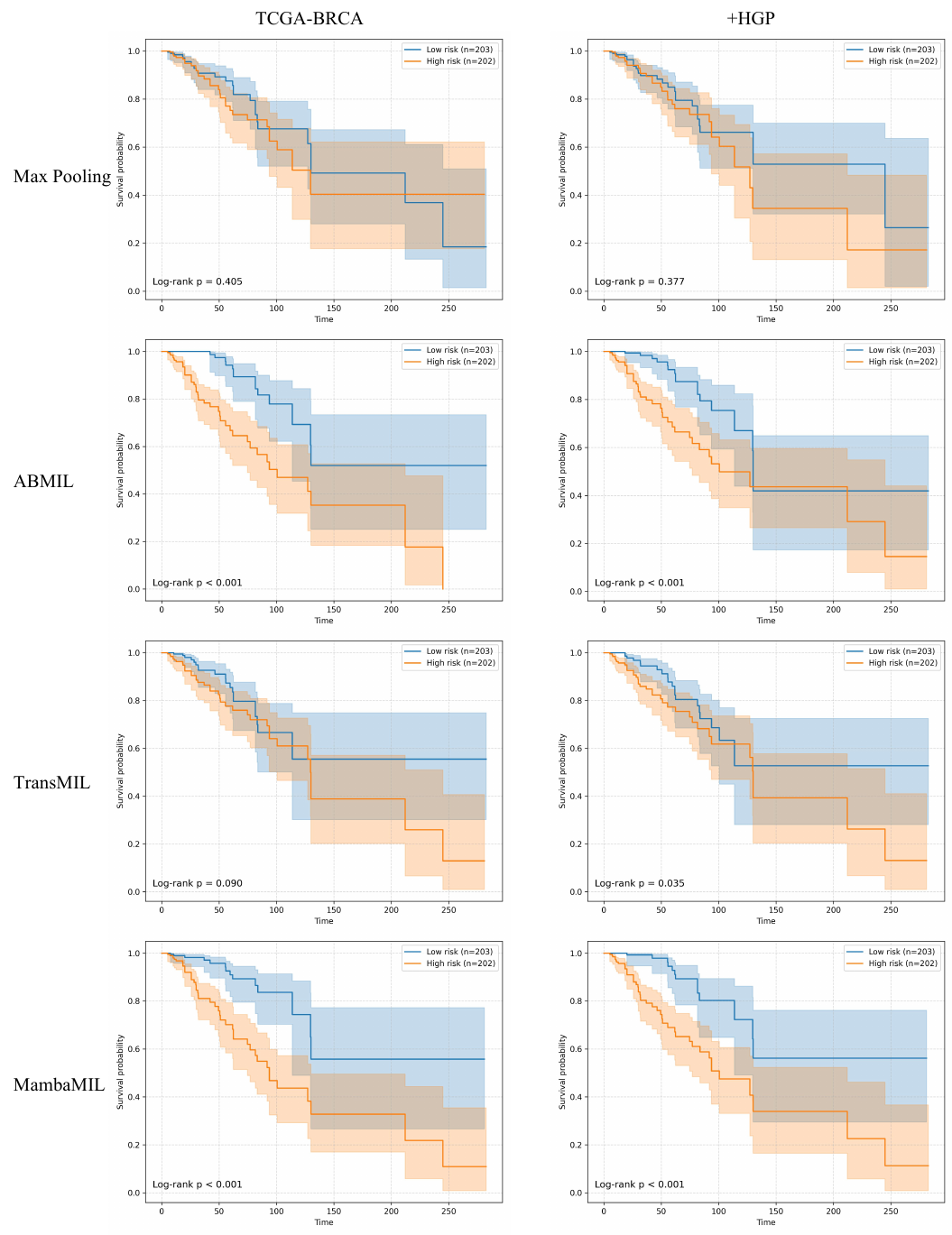}
    \caption{ Survival prediction results on the TCGA-BRCA dataset.}
    \label{fig:survival}
\end{figure}

Survival analysis aims to assess the model’s ability to distinguish patients with different survival outcomes based on histopathological features. As illustrated in Fig.~\ref{fig:survival} (the Kaplan–Meier curves), our method demonstrates substantially superior survival prediction performance compared to other approaches. 

In contrast to the original MIL variants, the +HGP models exhibit clearer separation between high- and low-risk groups, with substantially lower p-values. These findings highlight that our model not only improves classification accuracy but also generalizes effectively to complex prognostic tasks. By leveraging histomorphology prior, HGPMIL captures biologically meaningful representations associated with patient survival, providing a robust foundation for clinical decision support in computational pathology.

\begin{figure}[t]
    \centering
    \includegraphics[width=1\linewidth]{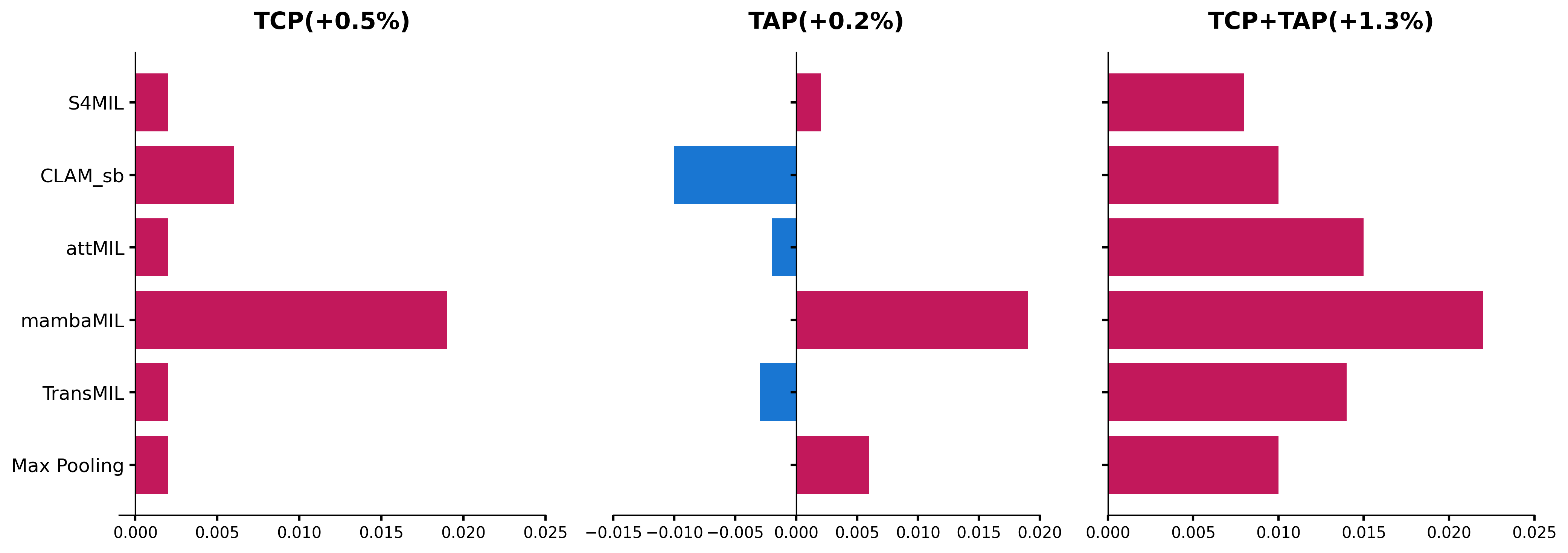}
    \caption{Ablation experiment results on AUC change.}
    \label{fig:ablation_auc}
\end{figure}

\begin{figure}[t]
    \centering
    \includegraphics[width=1\linewidth]{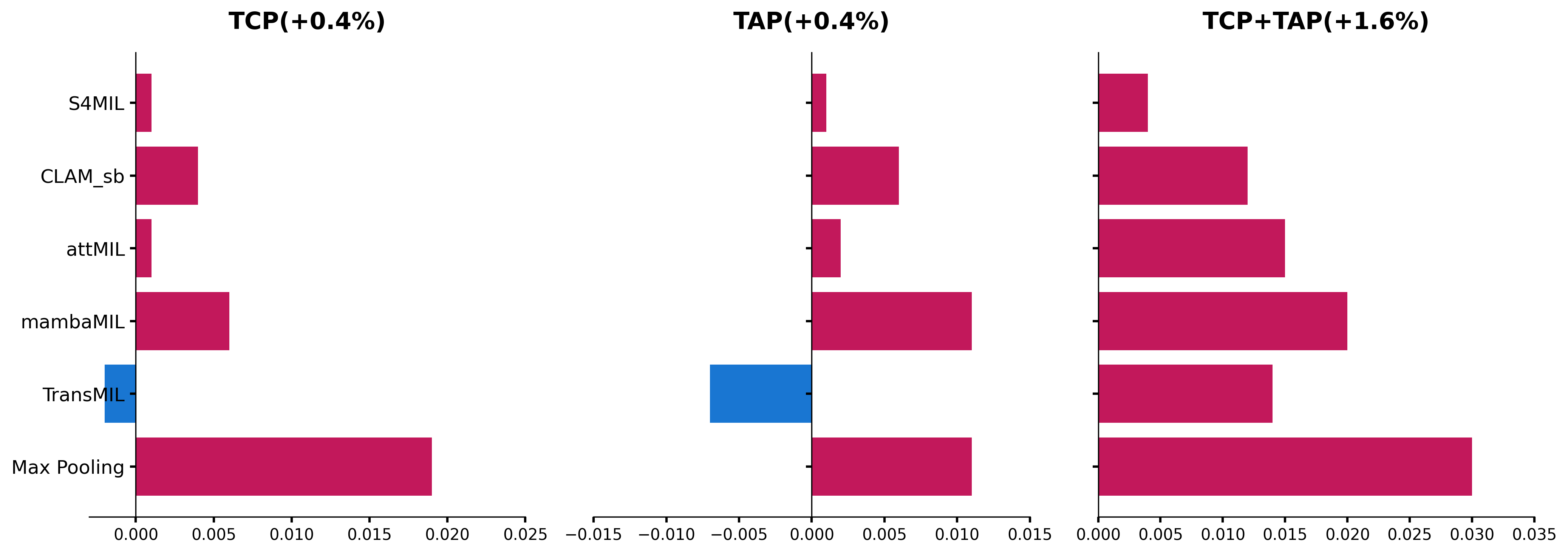}
    \caption{Ablation experiment results on ACC change.}
    \label{fig:ablation_acc}
\end{figure}

\begin{figure}[!htbp]
    \centering
    \includegraphics[width=1\linewidth]{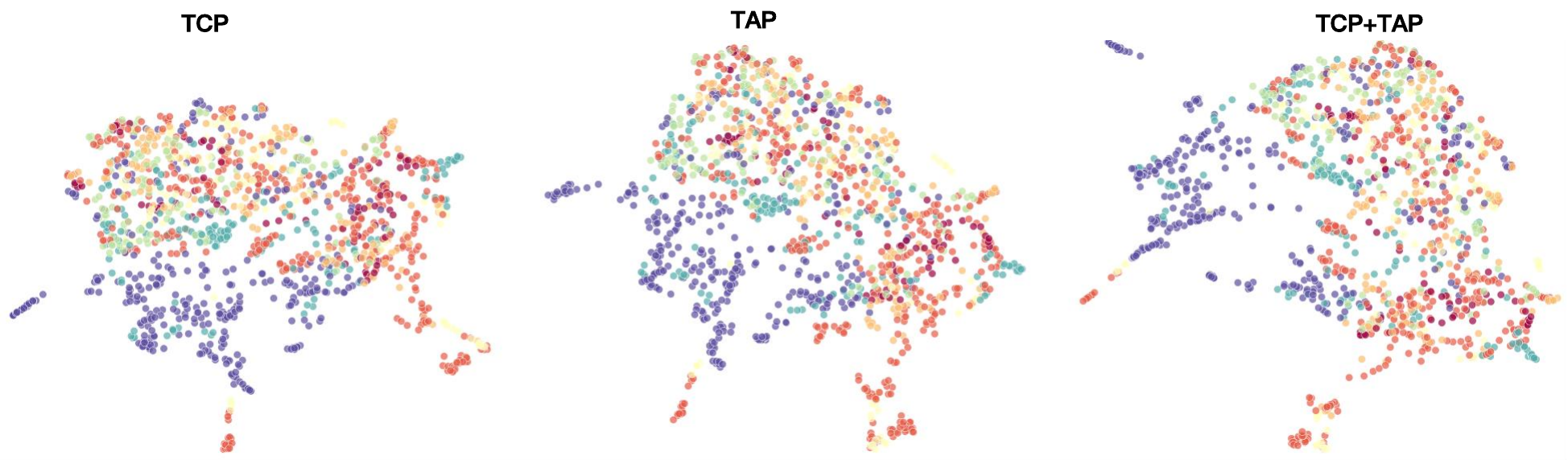}
    \caption{UMAP visualization for BRACS, each color represents a tumor subtype.}
    \label{fig3}
\end{figure}

\subsection{Ablation study}

To evaluate the effectiveness of HGPMIL, we perform an ablation study comparing the performance of its two key components: tumor cellularity prototypes (TCP) and tumor tissue architecture Prototypes (TAP). To ensure a fair evaluation of each component, we maintain identical training settings across all experiments.

As illustrated in Fig.~\ref{fig:ablation_auc}, TCP improves AUC by an average of 0.5\% across MIL methods on the BCNB dataset, while TAP contributes a 0.2\% gain. Similarly, Fig.~\ref{fig:ablation_acc} shows that on the TCGA-BRCA dataset, TCP and TAP each improve ACC by 0.4\% on average. When combined, HGPMIL yields a more notable performance boost, with average gains of 1.3\% in AUC (BCNB) and 1.6\% in ACC (TCGA-BRCA). These results confirm that both TCP and TAP effectively enhance molecular subtyping in MIL frameworks.

The Fig.~\ref{fig3} presents UMAP visualizations of slide-level embeddings on the BRACS dataset, color-coded by tumor subtypes. The histomorphological prototypes learned by HGPMIL form compact and distinct clusters in the feature space, indicating that prototype-guided learning enhances inter-class separation. Additionally, Fig.~\ref{fig:map} illustrate the distinct distributions of histomorphology scores and attention weights. These maps represent different modalities of the WSI, integrating histomorphological information. By combining these complementary features, HGPMIL achieves a deeper understanding of the WSI, guiding the model to focus on more relevant regions.

\begin{figure}[t]
    \centering
    \includegraphics[width=\linewidth]{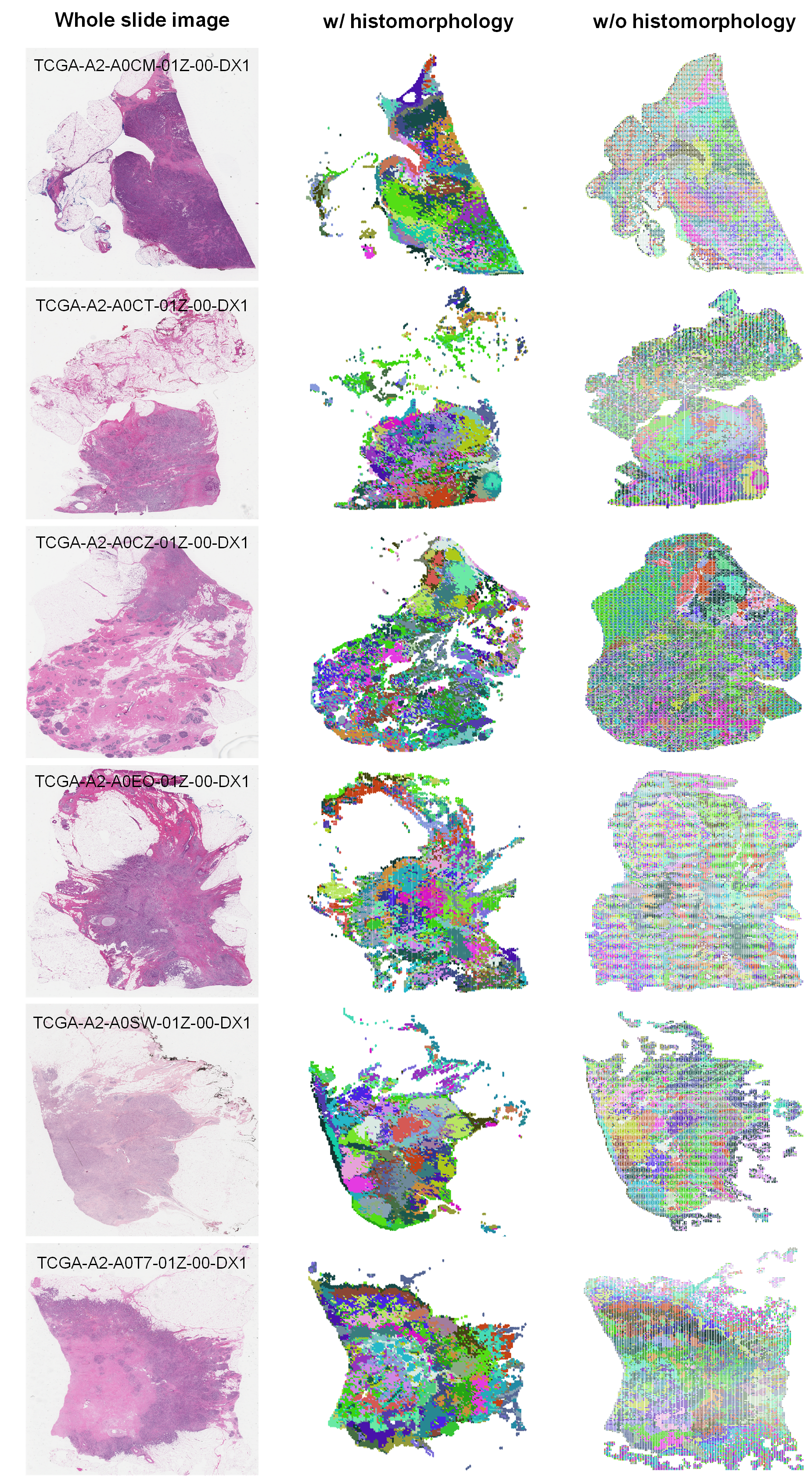}
    \caption{Clustering results visualization.}
    \label{fig:cluster}
\end{figure}


\begin{figure}[tbp]
    \centering
    \begin{subfigure}{0.2\textwidth}
        \centering
        \includegraphics[width=\linewidth]{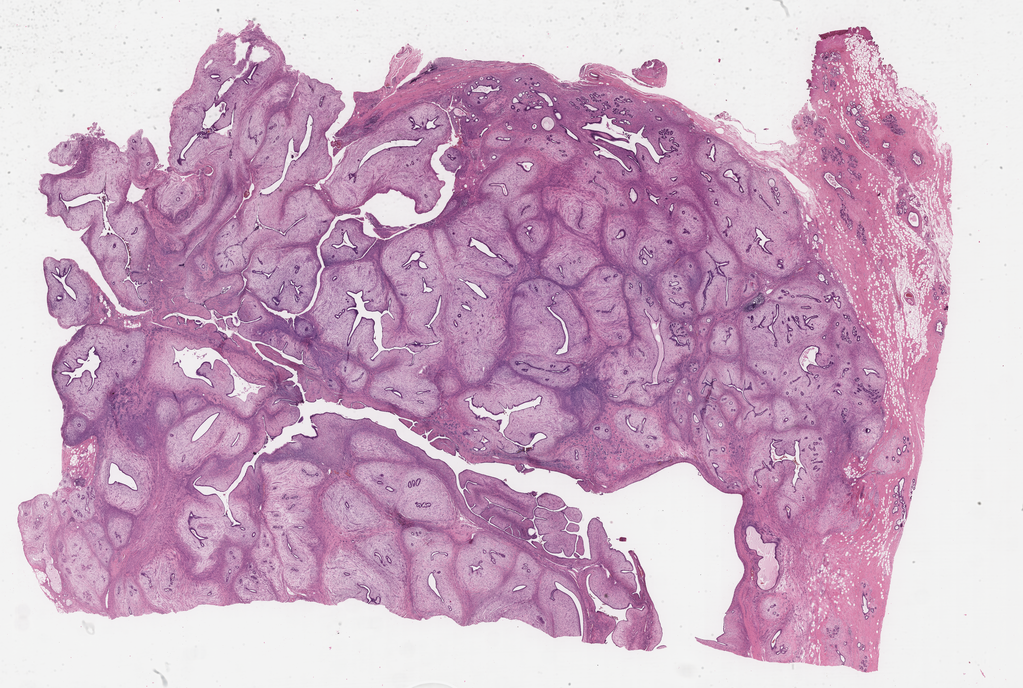}
        \caption{\footnotesize{H\&E slide}}
    \end{subfigure}
    \hfill
    \begin{subfigure}{0.2\textwidth}
        \centering
        \includegraphics[width=\linewidth]{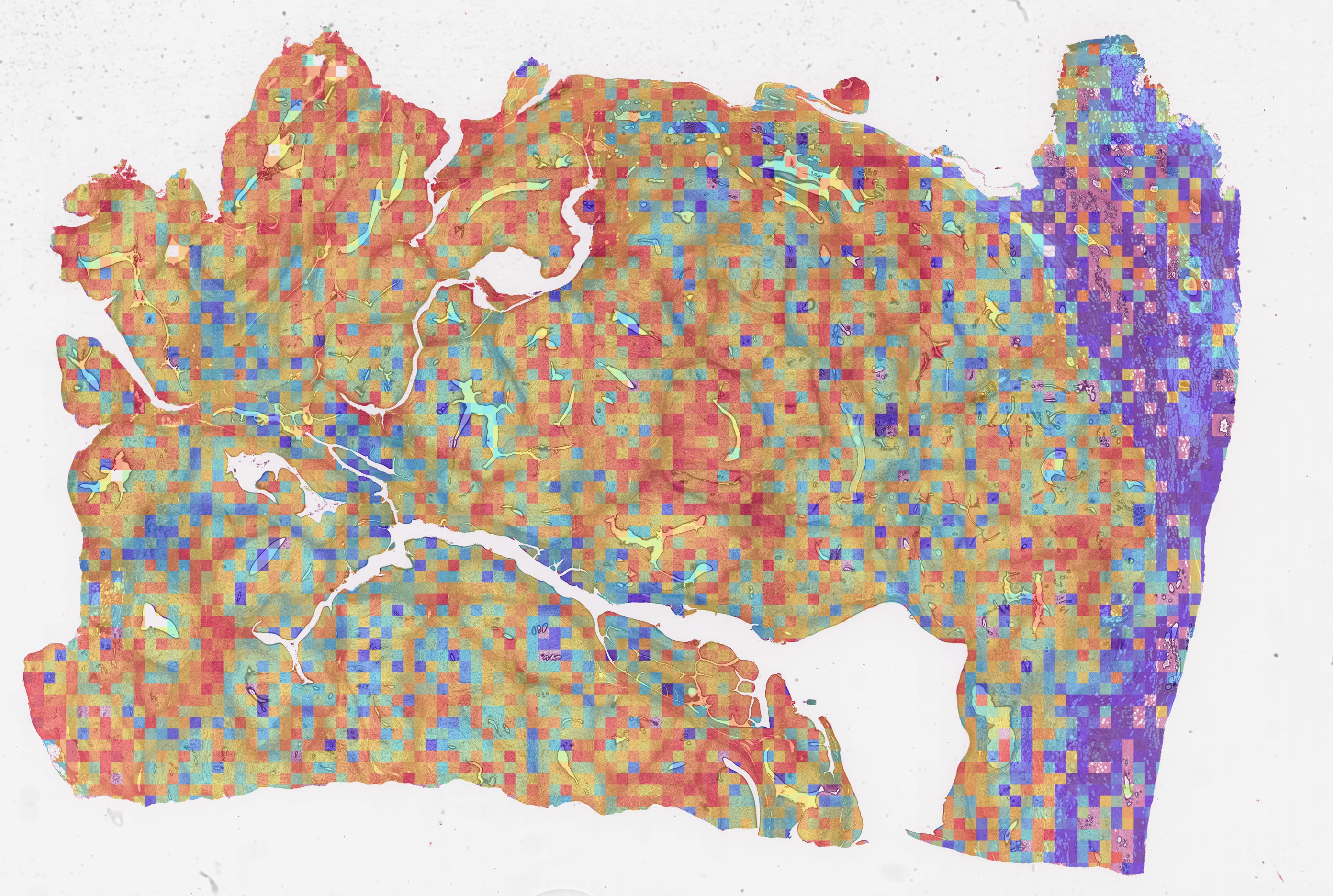}
        \caption{\footnotesize{Cellularity}}
    \end{subfigure}
    \hfill
    \begin{subfigure}{0.2\textwidth}
        \centering
        \includegraphics[width=\linewidth]{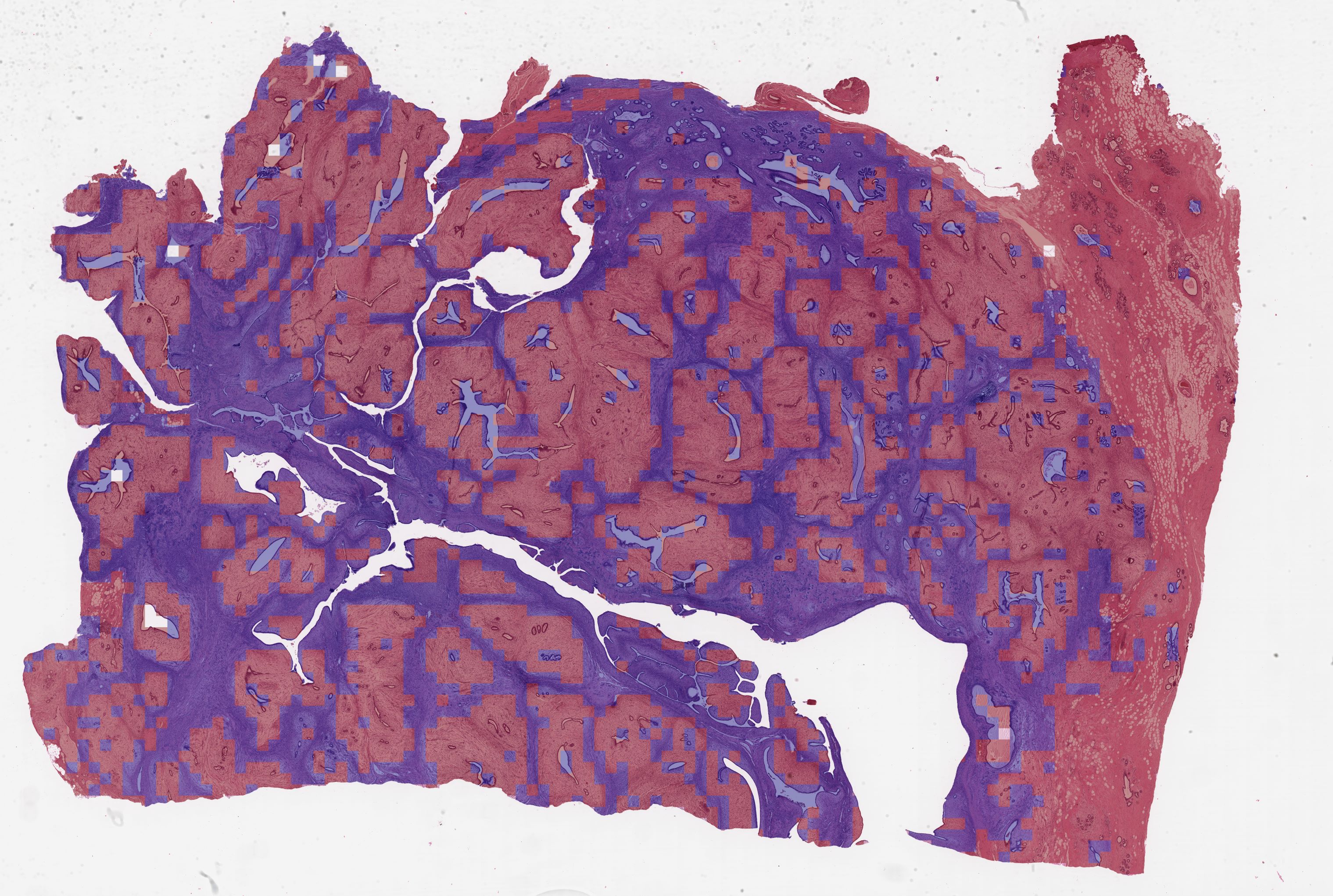}
        \caption{\footnotesize{Architecture}}
    \end{subfigure}
    \hfill
    \begin{subfigure}{0.2\textwidth}
        \centering
        \includegraphics[width=\linewidth]{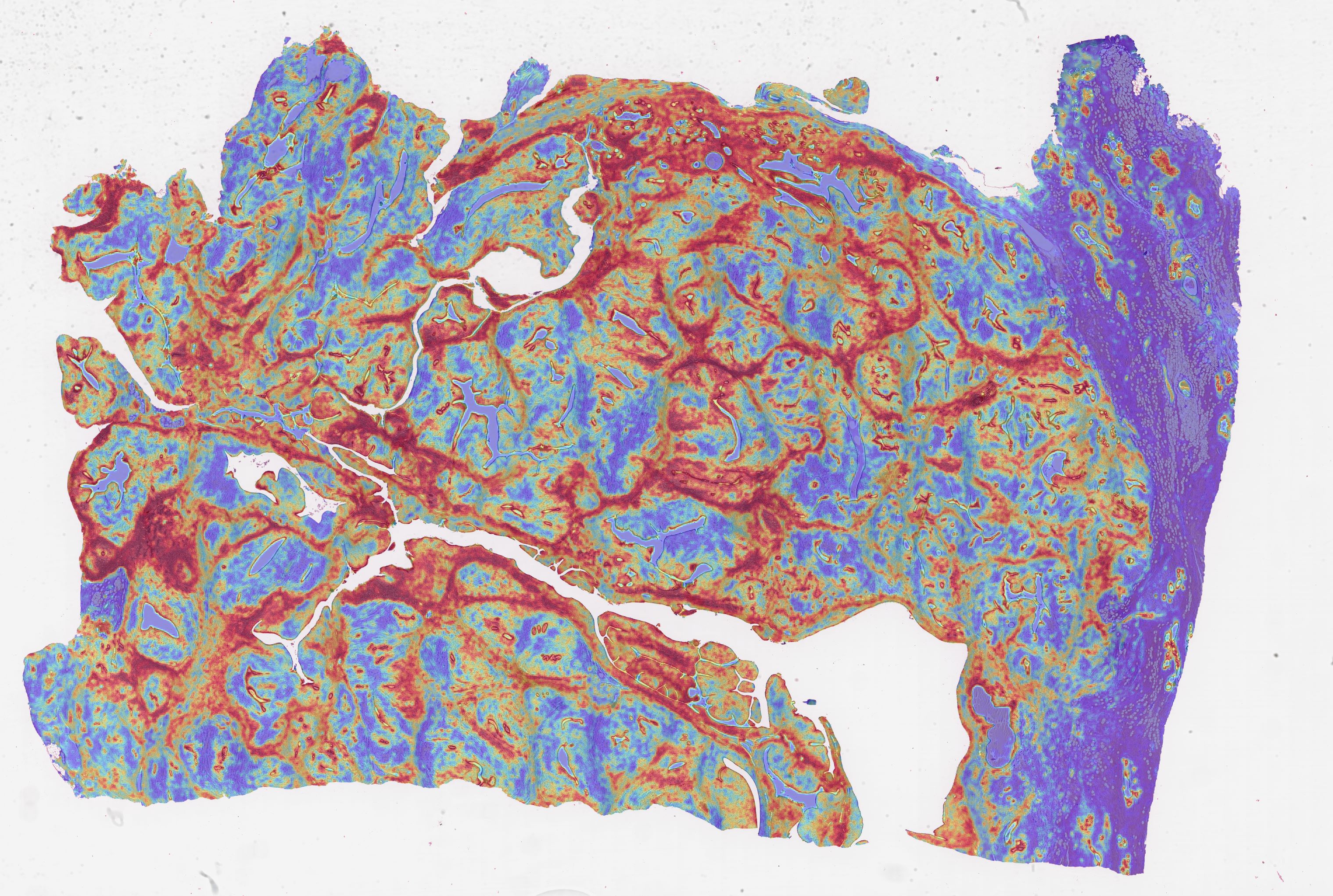}
        \caption{\footnotesize{Attention}}
    \end{subfigure}

    \begin{subfigure}{0.2\textwidth}
        \centering
        \includegraphics[width=\linewidth]{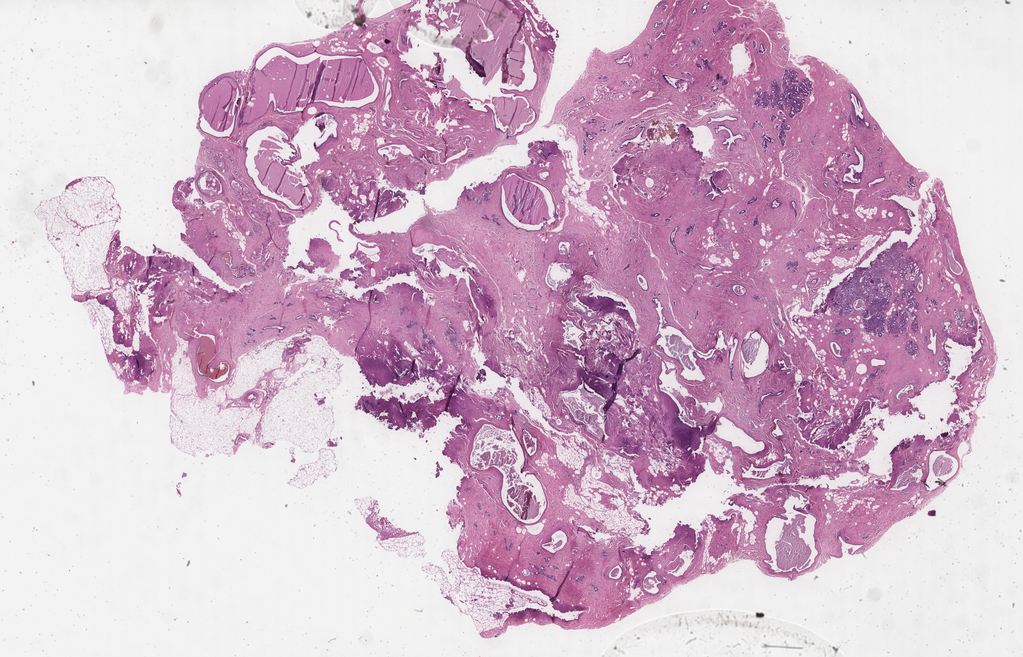}
        \caption{\footnotesize{H\&E slide}}
    \end{subfigure}
    \hfill
    \begin{subfigure}{0.2\textwidth}
        \centering
        \includegraphics[width=\linewidth]{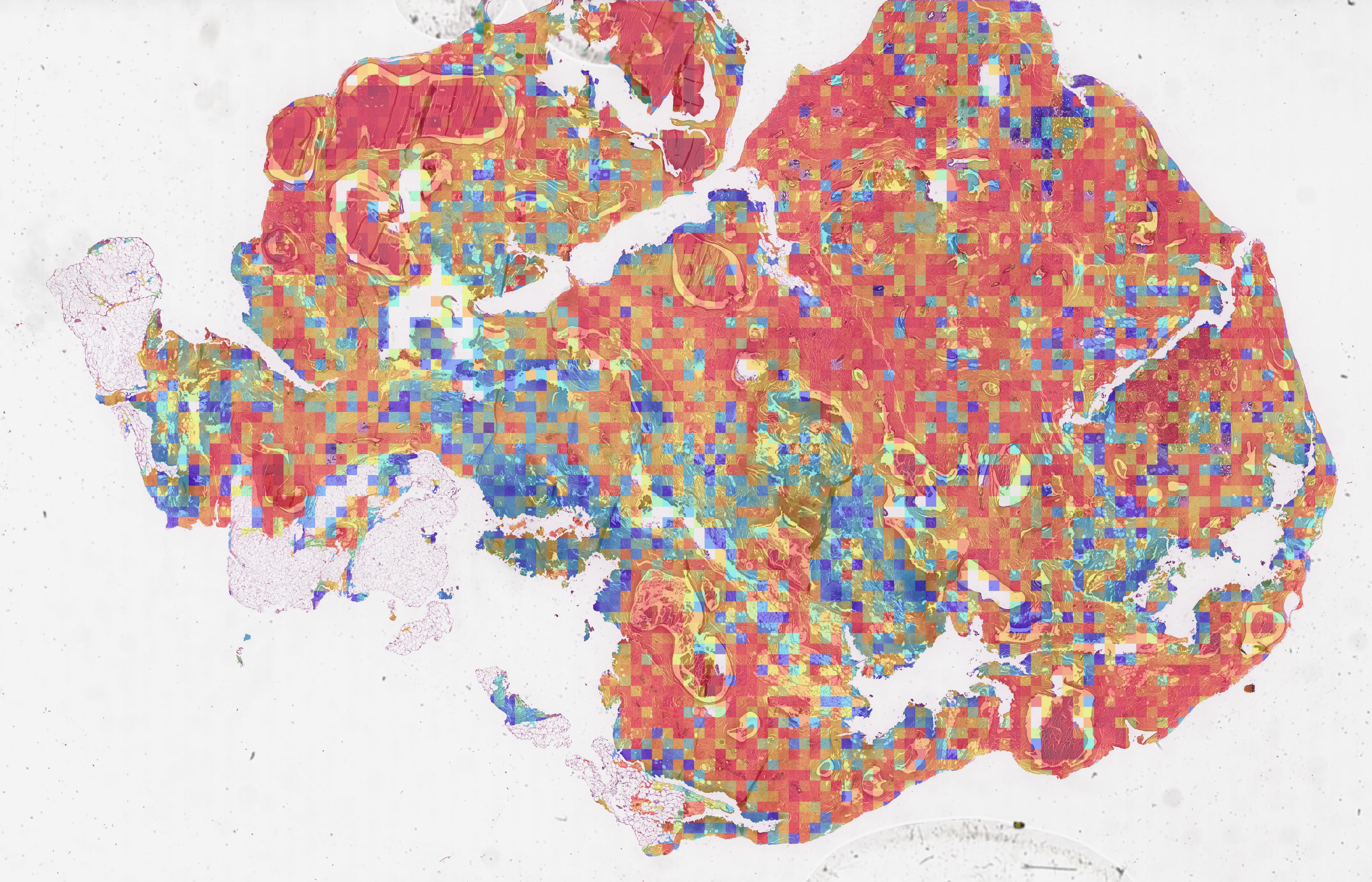}
        \caption{\footnotesize{Cellularity}}
    \end{subfigure}
    \hfill
    \begin{subfigure}{0.2\textwidth}
        \centering
        \includegraphics[width=\linewidth]{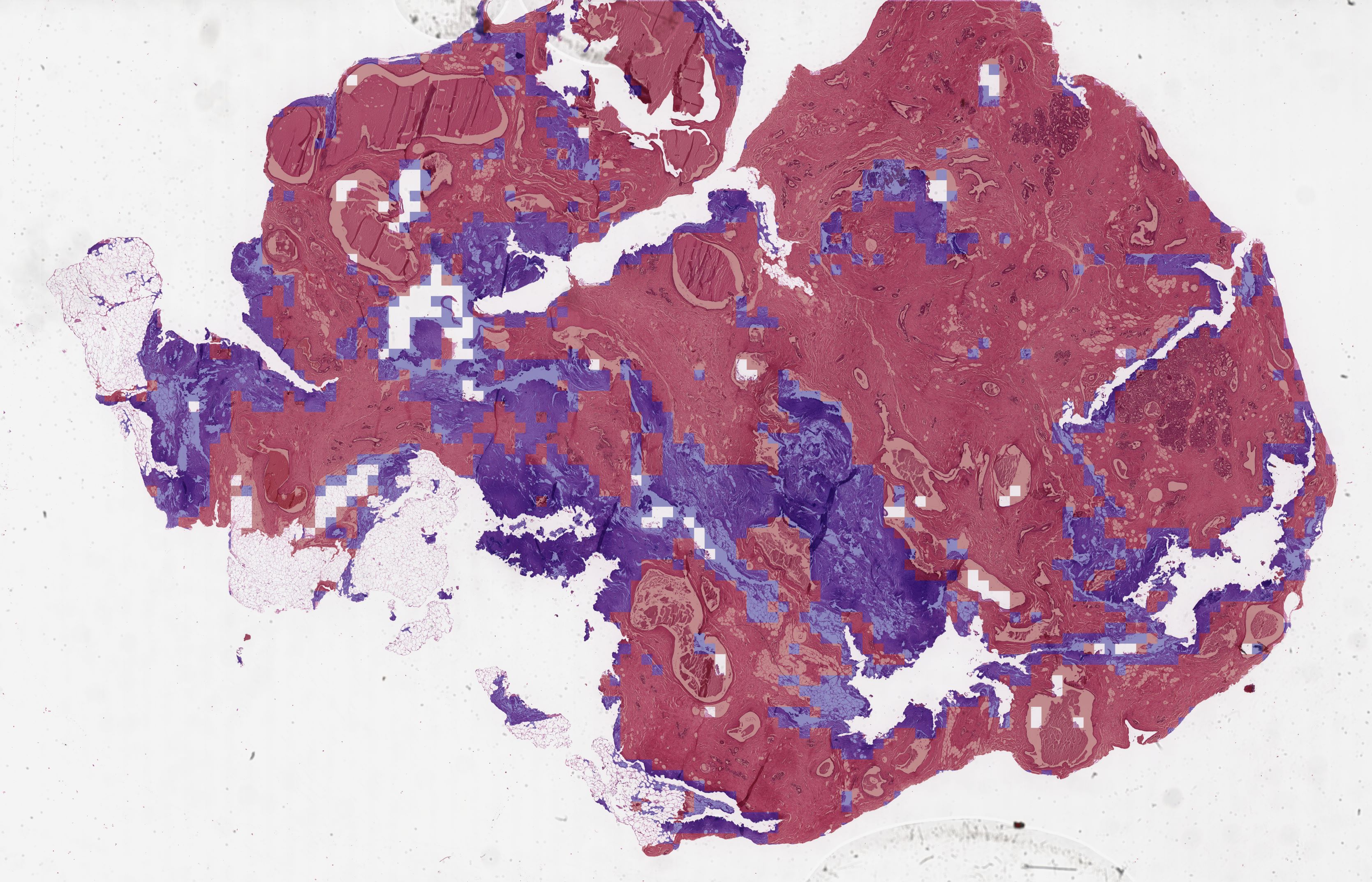}
        \caption{\footnotesize{Architecture}}
    \end{subfigure}
    \hfill
    \begin{subfigure}{0.2\textwidth}
        \centering
        \includegraphics[width=\linewidth]{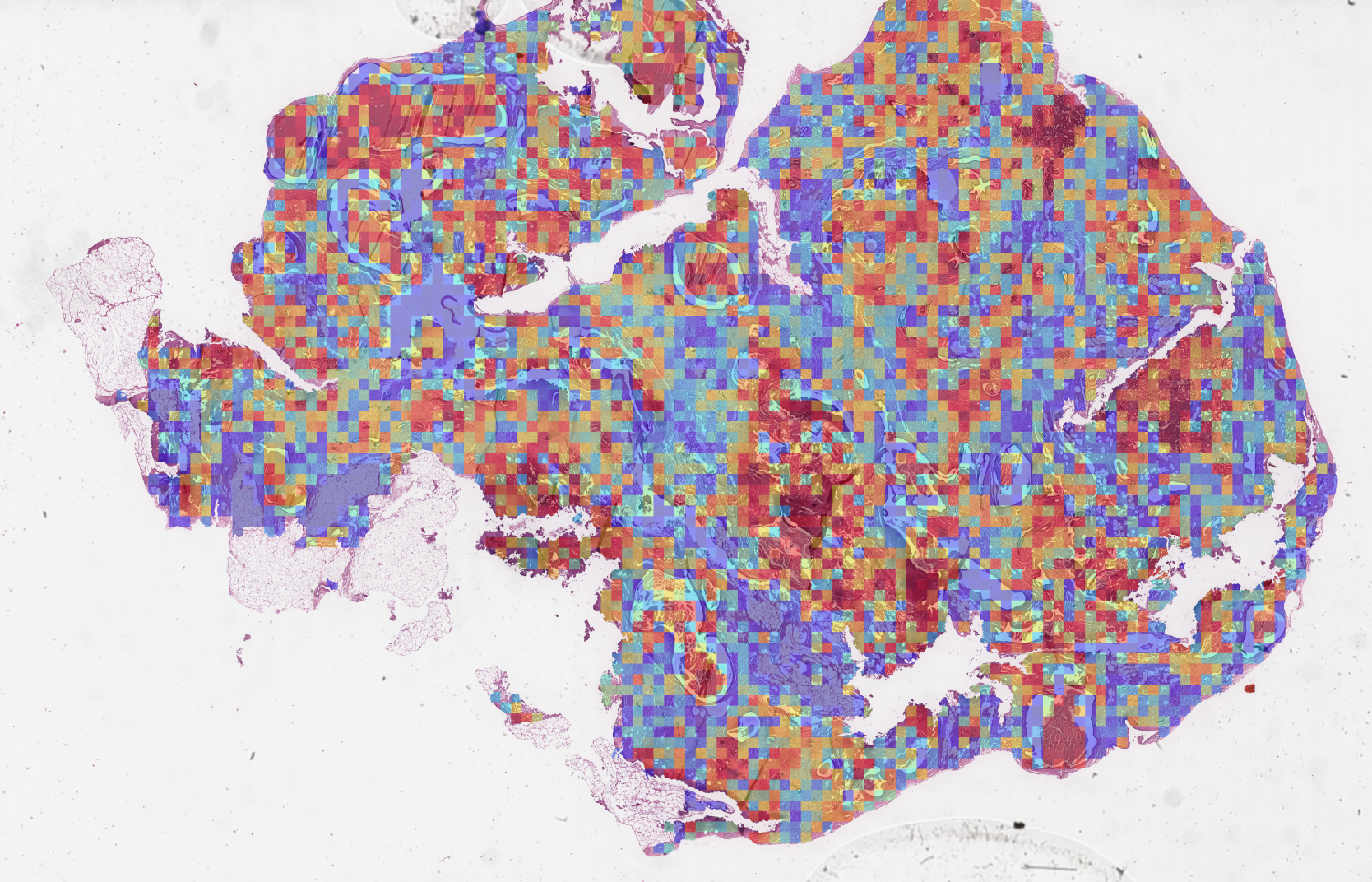}
        \caption{\footnotesize{Attention}}
    \end{subfigure}

\caption{Visualization of different importance scores. (a)(e) Original H\&E-stained slide. (b)(f) Tumor cellularity map. (c)(g) Tumor tissue architecture map. (d)(h) Attention weight map.}
    \label{fig:map}
\end{figure}

\begin{table}[!htbp]\small
\centering
\scriptsize  
\setlength{\tabcolsep}{3pt} 
\renewcommand{\arraystretch}{1.3}  
\caption{Performance comparison across diffferent cluster size on three datasets. \textbf{Bold} indicates the best result, and \underline{underline} denotes the second-best.}
\begin{tabular}{
    l
    |c c
    |c c
    |c c
    |c c
}
\hline
\rowcolor{gray!15}
\textbf{Cluster Size $K$} 
& \multicolumn{2}{c|}{\textbf{BCNB}} 
& \multicolumn{2}{c|}{\textbf{BRACS}} 
& \multicolumn{2}{c|}{\textbf{TCGA-BRCA}} 
& \multicolumn{2}{c}{\textbf{Mean}} \\
\hline
& AUC & ACC & AUC & ACC & AUC & ACC & AUC & ACC \\
\hline
$K$=10  & 0.816 & \textbf{0.797} & \underline{0.863} & \underline{0.633} & \textbf{0.856} & \textbf{0.836} & \underline{0.845} & \underline{0.755}\\
$K$=20  & 0.819 & 0.791 & 0.860 & 0.631 & 0.846 & 0.828 & 0.841 & 0.750\\
$K$=50  & \textbf{0.826} & 0.792 & \textbf{0.869} & \textbf{0.650} & \underline{0.855} & \underline{0.832} & \textbf{0.850} & \textbf{0.758}\\
$K$=75  & \underline{0.820} & \underline{0.793} & 0.862 & 0.624 & 0.838 & \textbf{0.836} & 0.840 & 0.751\\
$K$=100 & 0.819 & 0.792 & 0.854 & 0.629 & 0.844 & 0.830 & 0.839 & 0.750\\
\hline
\end{tabular}
\label{tab:cluster_ablation}
\end{table}
We compared our proposed histomorphology-prototypical clustering with traditional K-means clustering. As shown in Fig.~\ref{fig:cluster}, by incorporating histomorphological knowledge, histomorphology-prototypical clustering better captures the key morphological features within the WSI, leading to more accurate clustering. In contrast to K-means, our model is able to effectively ignore irrelevant instances, focusing on the most diagnostically significant regions.
Varying the cluster size has a noticeable impact on model performance. We conducted an ablation study based on the ABMIL method across three datasets, as shown in Table~\ref{tab:cluster_ablation}. The best performance for BRACS, TCGA-BRCA, and BCNB datasets occurs at 50 clusters, with AUCs of 0.869, 0.856, and 0.826, respectively. TCGA-BRCA accuracy is stable at 10 and 75 clusters (both 0.836).
These results suggest that a moderate number of clusters (around 50) generally leads to robust and balanced performance. Too few clusters may miss discriminative patterns, while too many may overwhelm the attention mechanism, reducing its ability to focus on informative regions.

\begin{table}[!htbp]\small
\centering
\scriptsize  
\setlength{\tabcolsep}{3pt} 
\renewcommand{\arraystretch}{1.3}  
\caption{Performance comparison across different select ratio $R$.}
\begin{tabular}{
    l
    |c c
    |c c
    |c c
    |c c
}
\hline
\rowcolor{gray!15}
\textbf{Select Ratio} 
& \multicolumn{2}{c|}{\textbf{BCNB}} 
& \multicolumn{2}{c|}{\textbf{BRACS}} 
& \multicolumn{2}{c|}{\textbf{TCGA-BRCA}} 
& \multicolumn{2}{c}{\textbf{Mean}} \\
\hline
& AUC & ACC & AUC & ACC & AUC & ACC & AUC & ACC \\
\hline
$R$=10\% & \underline{0.825} & \textbf{0.793} & 0.862 & 0.639 & 0.844 & \underline{0.831} & \underline{0.844} & 0.754 \\
$R$=20\% & 0.824 & 0.791 & 0.849 & 0.626 & 0.841 & 0.830 & 0.838& 0.749 \\
$R$=30\% & 0.820 & 0.789 & \underline{0.865} & \textbf{0.652} & 0.843 & 0.828 & 0.843 & \underline{0.756} \\
$R$=40\% & 0.818 & 0.787 & 0.864 & 0.646 & \underline{0.846} & 0.830 & 0.843 & 0.754 \\
$R$=50\% & \textbf{0.826} & \underline{0.792} & \textbf{0.869} & \underline{0.650} & \textbf{0.855} & \textbf{0.832} & \textbf{0.850} & \textbf{0.758} \\
\hline
\end{tabular}
\label{tab:patch_ratio_ablation}
\end{table}
Finally, we evaluate the effect of different ratios, as shown in Table~\ref{tab:patch_ratio_ablation}. The 50\% selection ratio consistently achieves the best performance across BCNB, BRACS, and TCGA-BRCA, reaching the highest AUC (0.826, 0.869, and 0.855) and ACC (0.792, 0.650, and 0.832), respectively. Although lower ratios such as 10\% and 30\% yield competitive results on certain datasets, their performance is less stable. These findings suggest that a 50\% ratio strikes the best balance between discriminative information and noise, leading to robust and generalizable performance.

\section{Conclusions}
This paper introduces HGPMIL, a novel framework that integrates histomorphology-guided prototypical learning into MIL models. By explicitly leveraging histomorphology information, HGPMIL learns prototypical representations of tumor regions that guide the clustering of instance-level features. These learned prototypes are then aggregated to form discriminative cluster-level representations, which significantly enhance both model interpretability and diagnostic accuracy. Extensive experiments and ablation studies demonstrate that HGPMIL consistently outperforms traditional MIL methods and generalizes well across different MIL architectures and downstream tasks, establishing it as a versatile and powerful tool for WSI analysis. For future work, we plan to incorporate more controllable text report priors and explore alternative implementations for importance assessment, to further improve the WSI classification performance and interpretability.



{
    \clearpage
    \small
    \bibliographystyle{ieeenat_fullname}
    \bibliography{ref}
}


\end{document}